\def\year{2022}\relax
\documentclass[letterpaper]{article} 
\usepackage{aaai22}  
\usepackage{times}  
\usepackage{helvet}  
\usepackage{courier}  
\usepackage[hyphens]{url}  
\usepackage{graphicx} 
\usepackage{natbib}  
\usepackage{caption} 
\DeclareCaptionStyle{ruled}{labelfont=normalfont,labelsep=colon,strut=off} 
\usepackage[utf8]{inputenc} 
\usepackage{url}            
\usepackage{booktabs}       
\usepackage{amsfonts}       
\usepackage{nicefrac}       
\usepackage{microtype}      
\usepackage{xcolor}         
\usepackage{multirow}
\usepackage{amsmath,bm,amssymb}
\usepackage{graphicx}
\usepackage{subfig}
\usepackage{caption}
\usepackage{arydshln}
\usepackage{array}
\usepackage{amsthm}

\theoremstyle{definition}
\newtheorem{definition}{Definition}[section]
\newtheorem{interpretation}{Interpretation}[section]

\title{Revisiting Methods for Finding Influential Examples}
\author {
    Karthikeyan K\textsuperscript{\rm 1} \thanks{Work done while at University of Copenhagen}, 
    Anders Søgaard \textsuperscript{\rm 2}
}
\affiliations {
    \textsuperscript{\rm 1} Duke University\\
    \textsuperscript{\rm 2} University of Copenhagen \\
    karthikeyan.k@duke.edu, soegaard@di.ku.dk,
}
\begin{document}

\maketitle

\begin{abstract}
Several instance-based explainability methods for finding influential training examples for test-time decisions have been proposed recently, including Influence Functions, TraceIn, Representer Point Selection, Grad-Dot, and Grad-Cos. Typically these methods are evaluated using LOO influence (Cook's distance) as a gold standard, or using various heuristics. In this paper, we show that all of the above methods are unstable, i.e., extremely sensitive to initialization, ordering of the training data, and batch size. We suggest that this is a natural consequence of how in the literature, the influence of examples is assumed to be independent of model state and other examples -- and argue it is not. We show that LOO influence and heuristics are, as a result, poor metrics to measure the quality of instance-based explanations, and instead propose to evaluate such explanations by their ability to detect poisoning attacks. Further, we provide a simple, yet effective baseline to improve all of the above methods and show how it leads to very significant improvements on downstream tasks. 


\end{abstract}

\section{Introduction}


Finding influential examples is intuitively meaningful. Imagine you want to estimate the relationship between the height and weight of chairs based on a sample of chairs from a second hand furniture store. Say you find four chairs in the store of weights and heights: 3kg and 60cm, 8.5kg and 85cm, 9kg and 90cm, and 9.5kg and 95cm, and estimate the coefficient to be 5.1 with an intersect of around 44. You are curious about the influence of the outlier (the small $i$th chair), though, and want to know its influence on your estimate. You compute the {\em leave-one-out} (LOO) influence estimate simply by estimating based the remainder, changing your estimate of the coefficient to 10. If you evaluate this on a chair of 10kg and 100cm, you will predict 95cm with your original model and 100cm with your new one; you can now compute LOO distance ($\hat{y}_j-\hat{y}_{j(i)}=$5cm) or Cook's distance \citep{cook1977detection,Cook2011}:  $$\frac{\sum_{j=1}^n (\hat{y}_j-\hat{y}_{j(i)})^2}{f\epsilon^{\mathsf{mse}}}$$ with $f$ the number of features, leaving you with $(5^2/\epsilon^{\mathsf{mse}})=1$ to quantify the influence of the $i$th training instance (the small chair).

In machine learning, computing such influence scores serve two purposes: (a) interpretability, and (b) detecting harmful training instances, such as in the above example. While in simple cases, you can afford to compute LOO influence or Cook's distance, this is generally not feasible for most cases in machine learning. Computing LOO influence scores per inference in an online fashion would mean you would have to train and make inferences with $n$ models. Commonly, faster methods to compute influence scores, uses LOO influence as gold standard \citep{pmlr-v70-koh17a, basu2021influence}. 

Influence sketching \citep{Wojnowicz_2016} and influence functions \citep{pmlr-v70-koh17a} are general methods for computing influence scores of training instances per inference, but they both rely on expensive computations, prohibitive of application at even moderate scale \citep{molnar2019}. This has led researchers to explore methods that jointly output predictions and influence. Such methods are often referred to as {\em global} instance-based interpretability methods (rather than local methods) \citep{Carvalho2019MachineLI,molnar2019,das2020opportunities}, and they include TraceIn \citep{pruthi2020estimating}, and Representer Point Selection \citep{yeh2018representer}. In addition, two simpler heuristics, Grad-Dot and Grad-Cos \citep{NEURIPS2019c61f571d}, are commonly used as baselines \citep{pruthi-etal-2020-learning}.

\paragraph{Contributions} In this paper, we begin with the observation that, in the context of deep learning, most of the above methods are very unstable and produce wildly different influence estimates across different random seeds, data orderings, and batch sizes. Moreover, we show that the same holds true even for the proposed gold standard of LOO influence. This seems to undermine the foundations for using these methods for interpretability and for detecting poisonous training examples. For example, if influence scores are sensitive to random seeds, we have no guarantee that removing influential examples will change our predictions in predictable ways. Moreover, if our gold standard depends on seeds, ordering, or batch size, it does not provide us with an objective way of evaluating the above methods. This problem is amplified by the fact that existing evaluation heuristics \citep{hanawa2021evaluation} can also be very misleading (see \S4.4).

To remedy this, we revisit the definition of influence and argue that influence should be estimated relative to model states and data samples. If influence is estimated relative to model states and data samples, we need to sample from the likely model states to make inferences about the influence of data points relative to samples. In other words, in order for us to use influence scores to remove examples, say, to remove an unwanted bias, we need to do so not only based on actual influence scores computed for a single model, but based on {\em expected}~influence scores.  

We show that such {\em expected}~influence scores are much more stable across the above methods, and evaluate the impact of conditioning on model state on the downstream task of detecting poisonous attacks \citep{poison_svm,auditing_diff,gu2019badnets, chen2017targeted, chan-etal-2020-poison}. We find that {\em expected} TraceIn (whether exact or approximate) performs best on this task, across two datasets; influence functions, Representer Point Selection, and the Grad-Dot and Grad-Cos heuristics also benefit from computing expectations based on model state samples, achieving good results for one dataset, but fail to identify poisoning attacks for the other.

\section{Data and Methods}

\paragraph{Data} Throughout this work, we use the  MNIST\footnote{\url{https://keras.io/api/datasets/mnist/}. Distributed under a Creative Commons Attribution-Share Alike 3.0 license.} \citep{lecun2010mnist} and CoMNIST\footnote{\url{http://ds.gregvi.al/2017/02/28/CoMNIST/} \citep{comnist}. Distributed under a Creative Commons Attribyte/ShareAlike license.} handwriting recognition datasets. For MNIST, we only use examples labelled $0$ and $1$. We use a random, balanced sample of $1000$ training and $200$ test examples. For CoMNIST, we use examples with labels $A$ and $B$. We randomly sample $200$ balanced test examples, and use the rest of the examples for training (approx. $800$ examples). We use a simple CNN model with weight decay for training our classifiers. On both MNIST and CoMNIST, our classifiers exhibit accuracies in the 0.95-1.0 range. 

We briefly summarize the definition of LOO influence, as well as other influence methods that we evaluate below:  

\begin{definition}[Leave-one-out] To calculate the, $I(x_{train}, x_{test} )$, the influence of a training example, $x_{train}$, on a test example $x_{test}$, we first train the model using all the training examples and find the model's loss on $x_{test}$, call it $L_{D}(x_{test})$, then we remove $x_{train}$, and retrain the model using the same hyperparameters (including the same random initialisation). 
We then calculate the loss on $x_{test}$ using the retrained model, i.e., $L_{D/x_{train}}(x_{test})$. The difference between the two loss, $L_{D/x_{train}}(x_{test}) - L_{D}(x_{test})$ is the influence of $x_{train}$. \end{definition}

Influence functions (IFs) \citep{hampel86} are a core component of classic statistical theory and can be seen as approximations of leave one out (for convex functions), first used by \citet{pmlr-v70-koh17a} in the context of deep neural networks, and defined as: 

\begin{definition}[Influence functions]$I(x_{train}, x_{test} ) = \nabla_{\theta} L(x_{test}, \hat{\theta} )^{T} H_{\hat{\theta}}^{-1} \nabla_{\theta} L(x_{train}, \hat{\theta} )$, where $\hat{\theta}$ is the weights of the trained model and $H_{\hat{\theta}} = \frac{1}{n} \sum_{i=1}^{n} \nabla_{\theta}^{2} L(x_i, \hat{\theta}) $.\end{definition}

TraceInIdeal was introduced in \citet{pruthi2020estimating} and assumes we trained our model with a batch size of $1$. When we update the model weights from $W_t$ to $W_{t+1}$ because of a gradient step on $x_{train}$, we calculate the difference in loss on $x_{test}$ and attribute this change in loss to $x_train$. 
\begin{definition}[TraceInIdeal] In TraceInIdeal, $I(x_{train}, x_{test})$ id defined as the total change in loss of $x_{test}$ due to $x_{train}$.\end{definition}

TraceInCP \citep{pruthi2020estimating} is an approximation of TraceInIdeal and is applicable for models trained with any batch size. 

\begin{definition}[TraceinCP] $I(x_{train}, x_{test} ) = \sum_{i=1}^{k} \eta_i \nabla L(W_i, x_{train}). \nabla L(W_i, x_{test}) $ where $W_i$ is the weights of the $i^{th}$ checkpoint, and $\eta_i$ is the learning rate between the $i-1^{th}$ and $i^{th}$ checkpoint. \end{definition}

Inspired by the representer theorem \citep{scholkopf2001generalized}, Representer Point Selection (RSP) \citep{yeh2018representer} shows that, given a well trained model, the pre-activation predictions of a test example $\Phi(x_{test},  \hat{\theta} )$ can be linearly decomposed as $\sum_{i=1}^n \alpha_i k(x_{i}, x_{test} ) $, where $x_i$ is the $i^{th}$ training data. 

\begin{definition}[Representer Point Selection] Influence is defined as $$I(x_{train}, x_{test} ) = \frac{-1}{2\lambda n}\frac{\partial L(x_{train}, \hat{\theta}  ) }{\partial \Phi (x_{train}, \hat{\theta} ) }  f(x_{train})^T f(x_{test}) $$ where $f$ is the output of the last hidden layer, and $\Phi$ is the pre-activation output.\end{definition}

\begin{table*}[htb]
\centering\small
\begin{tabular}{ p{2.2cm}  p{1.2cm}  p{1.2cm}  p{1.2cm}   p{1.2cm}  p{1.2cm}  p{1.2cm}   }
  \toprule
     & \multicolumn{3}{c}{\textbf{MNIST}} & \multicolumn{3}{c}{\textbf{CoMNIST}} \\
     \cmidrule(lr){2-4}
    \cmidrule(lr){5-7}
    \textbf{Type} &  \textbf{Pearson} & \textbf{Spearman} & \textbf{90$^{\mathbf{th}}\cap$} & \textbf{Pearson} & \textbf{Spearman} &  \textbf{90$^{\mathbf{th}}\cap$} \\
    \midrule
    & \multicolumn{6}{c}{\textbf{Sensitivity to initialization}} \\
    \midrule
     \textbf{Leave-one-out} & 0.043 &  0.037 &  18.80 &  0.008 &  0.004 &  11.64 \\
     \midrule
     \textbf{IFs} & 0.023 &  0.014 &  26.08 & 0.000 &  0.004 &  26.82 \\
     \midrule 
     \textbf{TraceIn Ideal} & 0.574 &  0.693 &  55.31 &  0.351 &  0.650 &  45.23 \\
     \textbf{TraceIn CP} & 0.782 &  0.830 &  55.78 & 0.791 &  0.937 &  62.95 \\
     \textbf{RPS} &  0.862 &  0.906 &  62.48 & 0.628 &  0.897 &  63.74\\
     \textbf{Grad-Dot} & 0.782 &  0.664 &  59.10 & 0.617 &  0.909 &  63.39\\ 
    \textbf{Grad-Cos} &  0.955 &  0.867 &  55.42&  0.963 &  0.824 &  34.59\\
    \midrule
    & \multicolumn{6}{c}{\textbf{Sensitivity to ordering of training data}} \\
    \midrule
    \textbf{Leave-one-out} & 0.112 &  0.046 &  16.61 & 0.004 &  0.005 &  10.32 \\  
    \midrule 
    \textbf{IFs} & 0.071 &  0.066 &  32.77 & 0.005 &  0.005 &  31.65\\
    \midrule 
    \textbf{TraceInIdeal} & 0.354 &  0.432 &  19.04 & 0.210 &  0.497 &  22.30\\
    \textbf{TraceInCP} & 0.849 &  0.954 &  78.61 & 0.826 &  0.953 &  70.36\\
    \textbf{RPS} & 0.972 &  0.982 &  84.06 &  0.696 &  0.934 &  77.71\\
    \textbf{Grad-Dot} & 0.933 &  0.774 &  81.59 & 0.645 &  0.874 &  74.53\\  
    \textbf{Grad-Cos} &  0.976 &  0.956 &  71.24 & 0.931 &  0.882 &  50.26\\
    \midrule
    & \multicolumn{6}{c}{\textbf{Sensitivity to batch size}} \\
    \midrule
    \textbf{Leave-one-out} & 0.147 &  0.116 &  27.40 & 0.014 &  0.016 &  13.18 \\ 
    \midrule 
    \textbf{IFs} & 0.011 &  0.010 &  29.74 & 0.010 &  0.010 &  28.91\\
    \midrule 
    \textbf{TraceIn CP} & 0.739 &  0.938 &  75.28 & 0.819 &  0.943 &  68.33\\
    \textbf{RPS} & 0.936 &  0.980 &  80.19 & 0.693 &  0.904 &  69.50\\
    \textbf{Grad-Dot} & 0.901 &  0.704 &  76.47 & 0.676 &  0.930 &  71.04\\  
    \textbf{Grad-Cos} & 0.972 &  0.951 &  72.32 &  0.968 &  0.862 &  42.95\\
  \bottomrule\\
\end{tabular}
\caption{\label{tab:instability}\textbf{Sensitivity of influence methods to initialization, ordering, and batch size} 
For each of the 200 test example, we calculate the influence score for each of the training examples and find the average correlation metrics 
between the influence scores, across pairs of random seeds, randon orderings, or different batch sizes. 
Our results show that all influence methods are extremely sensitive to initialization, especially LOO influence and influence functions. The least sensitive methods -- RPS and the two heuristics -- are also the worst-performing methods in our downstream experiments below. }
\end{table*}

Finally, we present two baseline heuristics, i.e., relying on the dot products or cosine similarities between the gradients of the training and test examples: 

\begin{definition}[Grad-Dot] $I(x_{train}, x_{test} ) = \nabla_{\theta} L(x_{train}, \hat{\theta} )^T \nabla_{\theta} L(x_{test}, \hat{\theta} )$.\end{definition}

\begin{definition}[Grad-Cos] This baseline heuristic simply returns the cosine similarity between the gradients of train and test examples, i.e., $I(x_{train}, x_{test} ) = cosine(\nabla_{\theta} L(x_{train}, \hat{\theta} ),  \nabla_{\theta} L(x_{test}, \hat{\theta} ))$.\end{definition}


\section{Instability of Influence Methods}
Influence methods are used to explain model decisions \citep{pmlr-v70-koh17a}, detect mislabeled examples \citep{pruthi2020estimating}, or provide further information for humans working in human-in-the-loop settings \citep{feng2019ai}. These applications require that our influence scores are robust in the sense that the examples predicted to be mislabeled, were not simply harmful to this model, but are likely to be harmful to other models, as well. 



This section shows that the above methods are far from robust, but in fact {\em very} sensitive to random seeds, random orderings, and different batch sizes. We systematically study the instability of various methods for finding influential training examples by analysing their sensitivity to (i) random initialization; (ii) order in the which the training data is seen; and (iii) batch size. To study the sensitivity to random initialisation, for example, we fix all other hyperparameters (including the order in which the training data is seen) and vary only the seed for random initialization. 

For each of the test examples, we then compute influence score for all training examples relative to two models trained with different initializations, orderings or batch sizes. For such pairs of models, we calculate the correlation metrics (Pearson, Spearman, and the overlap between the 90th percentiles) between the two sets of influence scores. Using the 90th percentile overlap as an alternative metric to global correlations is motivated by the fact that we typically care most about the highly influential examples in real-life settings. Finally, we repeat each of these experiments with 4 different seeds (4C2 = 6 pairs) and present the average metrics. 





Note that deep neural networks are generally very sensitive to initialization \citep{frankle2018the}, so we expect some change in the influence scores. Influence scores nevertheless have to be relatively independent of initialization in order to be useful. If we retrain after removing predictably influential examples, we have no guarantee that this will change our model, if our influence scores are highly sensitive to initialization. We also observe that influence scores are very sensitive to random ordering and batch size; and possibly many other hyper-parameters. 

In Table~\ref{tab:instability}, we see that, among all the metrics, LOO influence and influence functions (IFs) are least stable. We calculate the exact influence function using the inverse of Hessian and observe that small changes in damping factor change the influence scores a lot. (The Hessian of a non-convex function is singular; hence, damping factor is required for inversion.) Note that the sensitivity of influence functions to damping factor has been studied previously \citep{basu2021influence}; but showing its (and other methods') sensitivity to initialization, ordering, and batch size is, to the best of our knowledge, a novel contribution. Note also how TraceInIdeal is more sensitive to random ordering than TraceInCP. This is expected, because TraceInIdeal is calculated by monitoring how the model changes after {\em each}~training step, whereas TraceInCP relies only on checkpoints after each epoch.

\section{Rethinking Influence}

Using LOO or Cook's distance as our gold standard, the influence of a training example is the effect of removing that data point on the output distribution. We argue that this definition only makes sense in the context of convex problems, where we can identify optimal solutions. In deep neural networks, where initializations, data ordering, and batch sizes can have large impact on our learned models, the definition becomes misleading, which is reflected by the instability of this gold standard itself. 

\subsection{Influence scores are non-trivially dependent on each other}\label{sec:infl_not-independent}


\begin{figure}
    \centering
    \includegraphics[scale=0.45]{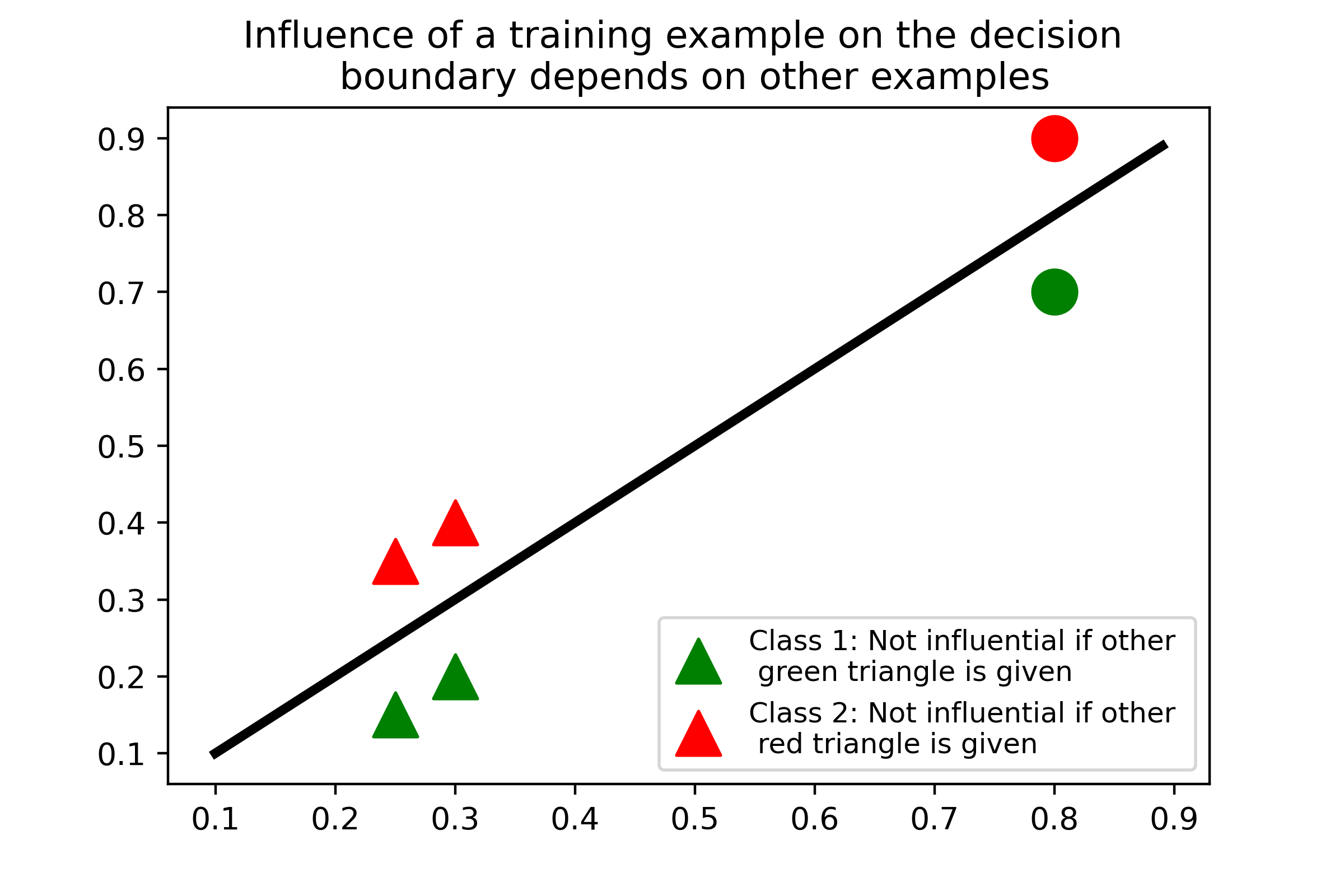} 
    \caption{Influence of a training data point depends on data distribution}
    \label{fig:data-dependent}
\end{figure}

Sometimes the way LOO or Cook's distance is used, suggests influence is a stable relation between training and test examples. One example is \citet{kocijan2020influence}, who use influence scores to resample data for transfer learning. They obtain negative results, which we believe are easily explained by the observation in the above that influence scores are unstable across initializations and reorderings. That is, there is no guarantee that the data with high influence scores is useful in the context of a new model, trained on a subset of the original data. 

Influence of a training data point depends on the data distribution even for convex problems. To illustrate this, consider Figure~\ref{fig:data-dependent}. If we remove only one of the green triangle (or red triangle), the decision boundary almost does not change, and hence the predictions on test examples also does not change much. It is misleading to say that the triangle data points are {\em not} influential, because removing both the green triangles (or both red triangles) completely change the decision boundary. In the context of non-convex problems where learning may be sensitive to random seeds and ordering, this problem is severely amplified. 
Note how TraceinIdeal to some extent captures this dependency. In our example, the green (or red) triangle that is seen earlier, might reduce more loss than the one that is seen later, and hence TraceinIdeal might consider, one of the green triangles to be more influential than the other; 
in contrast, methods like TraceinCP, RPS, Grad-Dot, Grad-Cos assign similar influence scores if the training examples are 'close'. 

\subsection{Influence scores are not independent on model state}\label{sec:model_state}


Deep neural networks are sensitive to random initialization \citep{frankle2018the}. If the model's prediction on a test example changes, then influence scores should also change. Influence scores in this sense depend on the final model state, but in another sense, the historical influence of a training example also depends on whether the example led to significant model updates in earlier iterations. 
Influence scores do not take this sensitivity into account, and the methods that are based on training checkpoints (TraceInIdeal and TraceInCP) do not sample from the space of model states, but rely solely on the states observed during training. 


Consider 
TraceinIdeal, for example, and where its instability arises. Assume a model is trained with a batch size of $1$ and SGD optimization. TraceinIdeal defines the influence of a training example $x_{train}$  on the test example $x_{test}$ as the total change in the loss on $x_{test}$ because of $x_{train}$. When we update the model from a state $W_t$ to $W_{t+1}$, due to a SGD step on example $x_{train}$, we measure the change in loss on the test example $x_{test}$. TraceinIdeal, gives the entire credit for the change in loss to $x_{train}$. However, the loss is a product, not only of $x_{train}$, but of $x_{train}$ and the current model state. Therefore, the change in loss on $x_{test}$ should be defined as influence of $x_{train}$ on $x_{test}$, given state $W_t$:

\begin{align*}
    &\textbf{TraceinIdeal Influence:} \\ &I(x_{train},x_{test})=L(x_{test}| W_{t})-L(x_{test} | W_{t+1})  \\
    &\textbf{Corrected  influence: } \\
&I(x_{train}, x_{test}| W_t) = L(x_{test}| W_{t}) - L(x_{test} | W_{t+1})\\
\end{align*}


In practice, we can approximate this by measuring the expected influence score over the model's state $E_{W_t}[I(x_{train}, x_{test}| W_t)]$. TraceinIdeal already does this to some extent, summing over a few states $W_t$. 
TraceInCP (Equation \ref{eqn:traceincp_grad}) is an  approximation of equation \ref{eqn:traceinCP_loss}: 

\begin{align}
    \begin{split}
        & TraceInCP(x_{train}, x_{test}) = \\
        &\qquad \sum_{i=1}^{k} L(x_{test}| W_i ) - L(x_{test} |W_{i+1}(x_{train}) ) \label{eqn:traceinCP_loss}
    \end{split}
    \\[2ex]
    \begin{split}
        &\qquad \approx \sum_{i=1}^{k}\eta \nabla_i l (W_{i}, x_{train}) . \nabla l (W_{i}, x_{test}) \label{eqn:traceincp_grad}
    \end{split}
\end{align}

We offer two interpretations of TraceInCP: 

\begin{interpretation}[TraceInCP]
If we define, $I(x_{train}, x_{test}| W_t)$, the influence of $x_{train}$ on $x_{test}$, at a given model state $W_t$, as the change in the loss on the $x_{test}$, if we had updated the model by taking a SGD step using $x_{train}$ -- $W_t$ to $W_{t+1}(x_{train}) $. Equation~\ref{eqn:traceinCP_loss} can then be seen as the state expected influence score, where we use checkpoints as the sample for $W_t$. 
\end{interpretation} 
In Interpretation~4.1, TraceInIdeal and TraceinCP differ only in the sampling of $W_t$.
\begin{interpretation}[TraceInCP]
 If we define $I(x_{train}, x_{test}| W_t)$ as the dot product between the gradients of $x_{train}$ and $x_{test}$ (Grad-Dot), Equation~\ref{eqn:traceincp_grad} can be seen as its expected influence score, where we use checkpoints as the sample for $W_t$, for calculating the empirical expectation.
 \end{interpretation}
 In Interpretation~4.2, TraceinCP coincides with {\em expected} Grad-Dot. 

\begin{figure}[htb]
    \centering
    \subfloat[Decision boundary with all training data]{\includegraphics[scale=0.45]{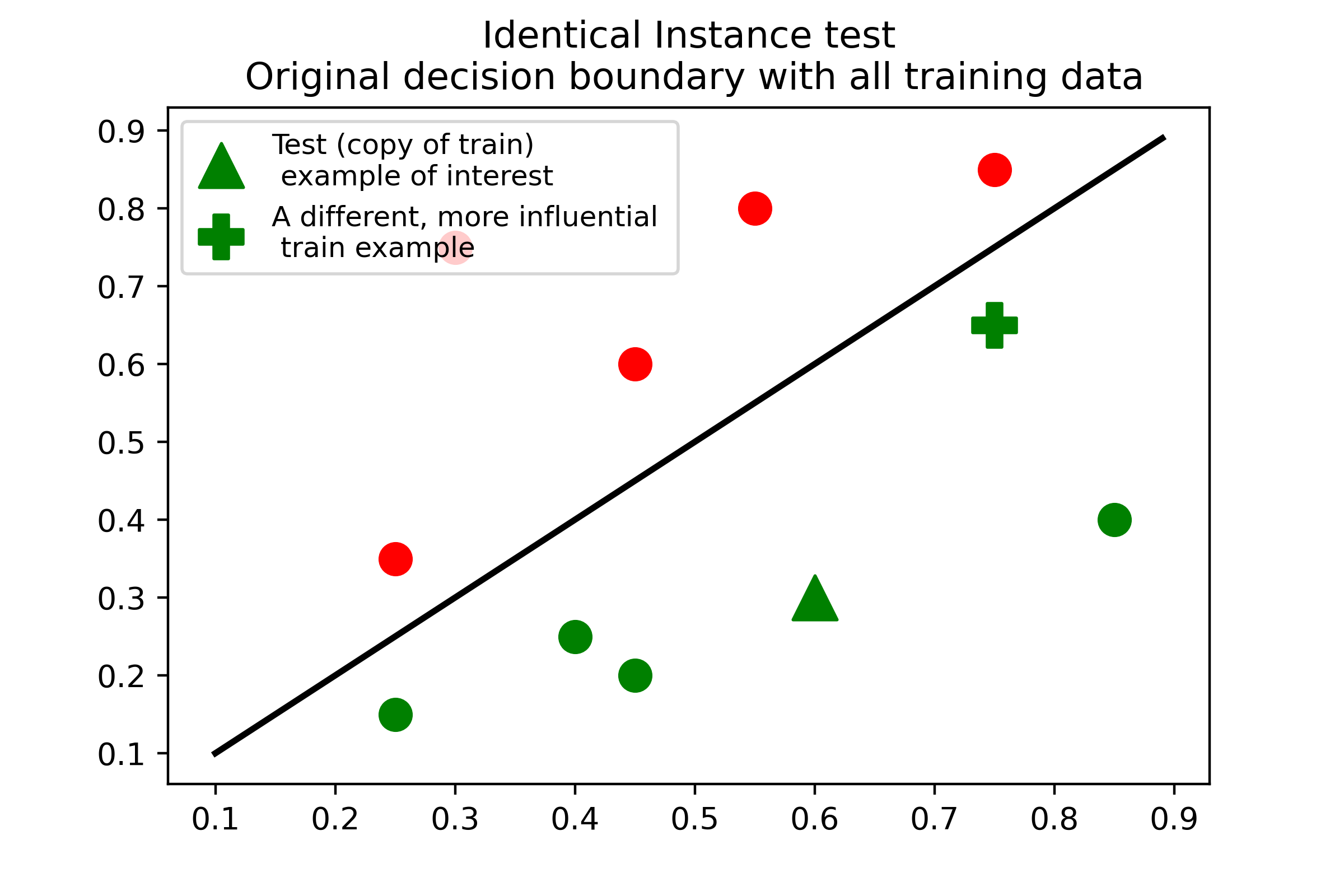}}
    
    \subfloat[Decision boundary with "+" removed]{\includegraphics[scale=0.45]{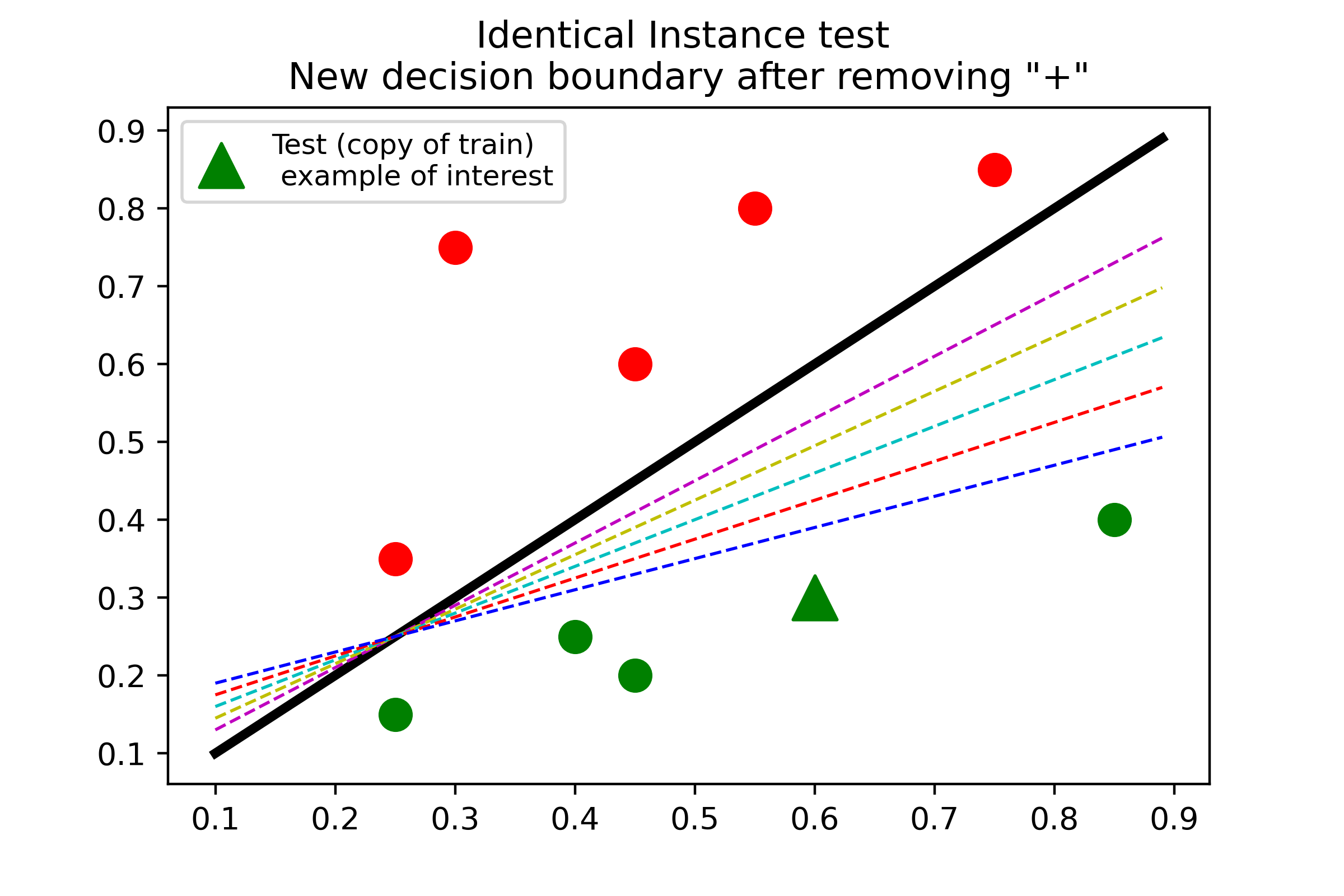}}
    \caption{\textbf{Conflicting example with Identical Instance Test:} Consider the green triangle as the point of interest. Removing it does not drastically affect the decision boundary, therefore the loss on triangle also does not change a lot.  However, removing "+" can change the decision boundary a lot and hence the loss on green triangle  can vary a lot. }
    \label{fig:identical_instance_test}
\end{figure}

\subsection{Why leave-one-out is therefore not a gold standard}\label{sec:loo-not-gold}

In LOO, the total reduction in the loss of $x_{test}$ is equal to the sum of the influence scores of all the training examples (Equation "original" in the Appendix). TraceIn also builds on this assumption. Our discussion in \S\ref{sec:model_state} shows how this assumption is false. The reduction in the loss of $x_{test}$ is the sum of the influence scores of the training examples only when conditioned on the state at which $x_{test}$ is seen. Assume that $x_{train}$ is first seen at time step $k$ in the original training phase. If we remove $x_{train}$ and retrain the model, every model state after $k$ changes; hence, the difference in loss is {\em not} equal to the influence of $x_{train}$. See Equation "corrected" in the Appendix for details.


\subsection{Why the Hanawa et al.~(2021) heuristics are also misleading}

To evaluate an influence metric, we need a gold standard to compare against, however, \S\ref{sec:loo-not-gold} and \S\ref{sec:infl_not-independent}, suggest we have to discard LOO as a gold standard. In the absence of a gold standard, we could turn to heuristics -- simple ways of quantifying properties we expect good influence scores to have. \citet{hanawa2021evaluation} proposes such heuristics to compare the quality of a influence metrics, including two criteria for sanity checks -- the identical instance test and the identical class test -- and two criteria for utility evaluation -- the top-$k$ identical class test and the identical subclass test. We argue that these heuristics are also misleading. 

\begin{definition}[The identical instance test] Assume the test example, $x_{test}$ is a copy of a train example $x_{train}$, and if the model predicts correctly on $x_{test}$, then the most influential training example should be $x_{train}$ itself. \end{definition}

Consider the green triangle in the figure~\ref{fig:identical_instance_test} to be the test example of interest (copy of train example). Note that removing it may not affect the decision boundary a lot. Therefore, the presence or absence of it in the training data does not drastically change the prediction (loss) on the test example (itself). However, if we remove the green "+", it can change the decision boundary a lot and hence the prediction (loss) or confidence in prediction, on the green triangle can also change a lot. Intuitively, the green "+" is more influential on the green triangle than the green triangle itself. The argument here is similar to that for SVM-like models -- support vectors are generally more influential than the data points away from the decision boundary or hyperplane. In general, the identical instance test assumption is only valid when the model memorises all the training examples; however, the neural networks (or parametric models) have a limited memory and relies only on the most informative patterns. 

\begin{definition}[Identical class test] If the model predicts class $y$ on the test example $x_{test}$, then the most influential training example should belong to class $y$. \end{definition} 

To predict examples as of class A, however, examples of class B can help in providing contrastive evidence. To see this, consider one-versus-all classification problems, e.g., identifying cats in images or all strings containing the letter {\em a}. Here, it is the consistent association of a pattern with one class (A) that is important also for classifying members of the other class (B). Members of class A and B are equally important for deriving this rule. 


The utility heuristics, the top-$k$ identical class test and identical subclass test, which are simple modifications of Definition~4.1 and 4.2 are misleading for similar reasons to what has been discussed.

\begin{table*}
\centering\small 
\begin{tabular}{ l l cccccc}
  \toprule
     & & \multicolumn{3}{c}{\textbf{MNIST}} & \multicolumn{3}{c}{\textbf{CoMNIST}} \\
      \cmidrule(lr){3-5}
    \cmidrule(lr){6-8}
    \multicolumn{2}{l}{\textbf{Method}} &  \textbf{Pearson} & \textbf{Spearman} & \textbf{90$^{\mathbf{th}}\cap$} & \textbf{Pearson} & \textbf{Spearman} &  \textbf{90$^{\mathbf{th}}\cap$} \\
  
    \midrule
    \multirow{7}{*}{\rotatebox[origin=c]{90}{\sc baseline}}&\textbf{LeaveOneOut} &  0.238  &  0.050 & 17.46 & 0.004  &  0.001 & 10.30  \\
              \cmidrule{2-8}
&\textbf{Influence Fn.} & 0.008  &  0.035 & 28.86 & 0.001  &  0.003 & 25.32 \\
              \cmidrule{2-8}
&\textbf{TraceIn Ideal} &  0.411  &  0.485 & 21.88 & 0.241  &  0.566 & 23.19  \\
     &\textbf{TraceIn CP} & 0.811  &  0.746 & 62.43 & 0.710  &  0.931 & 61.19  \\
     &\textbf{RPS} &  0.896  &  0.935 & 67.54 &  0.500  &  0.857 & 62.24 \\
     &\textbf{Grad-Dot} &  0.759  &  0.604 & 57.83 & 0.532  &  0.937 & 64.53  \\ 
    &\textbf{Grad-Cos} & 0.929  &  0.896 & 62.18 & 0.988  &  0.834 & 35.18 \\
      \midrule
         \multirow{7}{*}{\rotatebox[origin=c]{90}{\sc expected}}&\textbf{E-LeaveOneOut} & 0.595  &  0.130 & 25.83 & 0.007  &  -0.009 & 9.75 \\
         \cmidrule{2-8}
     &\textbf{E-Influence Fn.} &  0.005  &  -0.004 & 28.90 & -0.011  &  0.003 & 26.69   \\
              \cmidrule{2-8}
&\textbf{E-TraceIn Ideal} & 0.667  &  0.709 & 18.60 &  0.820  &  0.873 & 38.44 \\
     &\textbf{E-TraceIn CP} &   0.939  &  0.936 & 81.72 & 0.944  &  0.988 & 82.99 \\
     &\textbf{E-RPS} &   0.995  &  0.995 & 93.46 & 0.943  &  0.939 & 82.54  \\
     &\textbf{E-Grad-Dot} &  0.960  &  0.749 & 87.26 & 0.865  &  0.981 & 77.91 \\ 
    &\textbf{E-Grad-Cos} & 0.996  &  0.991 & 89.40 &  1.000  &  0.953 & 62.08 \\
  \bottomrule\\
\end{tabular}
\caption{\label{tab:marginal}\textbf{Expected influence metrics: } For each of the 200 test point, we calculate the correlation coefficients, as well as the overlap between the 90th percentile (\textbf{90$^{\mathbf{th}}\cap$}), between the expected influence scores of training examples, calculated using 2 different (and disjoint) sets of seeds. We report the average correlation coefficients over test examples.}
\end{table*}

\section{How to Improve Influence Methods}\label{sec:expected_influence}


As we have seen before, that influence of training example (on test example) can change based on the state of the model (section~\ref{sec:model_state}). However, the conditional (on state) influence scores may not be a reliable metric, a better metric would be to find expected influence score (expectation over over state). 


To find the empirical expectation, we need to sample $W_t$ from its distribution -- the probability of seeing $W_t$ as a state while training the model (possibly with different hyperparameters, initialization, etc..). To imitate sampling from the distribution of $W_t$, we train the model with different initializations and random ordering and use them to calculate the expedited influence score \footnote{we understand this may not be an unbiased sampling, and we could have changed other hyperparameters as well, but just for simplicity we fixed them}. For each test example, we calculate the influence scores of all the training examples using many different models (each with a different initialization and random ordering of training data) and calculate the expected influence as the average of the influence scores (averaged over different models).

Note that in section \ref{sec:model_state}, we have seen that TraceinIdeal and TraceinCP are already sum of conditional influence over multiple states. By averaging over states from different models we expect a better estimate. For RPS, Grad-Cos, Grad-Dot the influence scores depends only on the final state, we assume these scores to be the influence conditioned on the state (final state), and averaging over multiple models (their final states) gives expected influence. Although, final states may not be the best way to sample $W_t$, RPS, Grad-Dot and Grad-Cos is valid only at the final state (by their definitions). Also, Grad-Dot averaged over intermediate state is simply TraceinCP. From sec\ref{sec:loo-not-gold}, we can see that leave-one-out does not give the influence of $x_{train}$ conditioned on a state, however, expectation of influence scores calculated using leave-out-out metric would give us expected influence of the $x_{train}$ (refer appendix -- take expectation on both sides of equation "corrected", in appendix, and the terms not related to $x_{train}$ becomes zero). As there are many variables in the equation, finding this expectation could be noisy, especially, if we use only a small sample. Finally, as influence function is approximation of leave-one-out (atleast in theory), we assume averaging over multiple models would also give expected influence.  

To measure the stability of expected influence, we calculate the expected influence using 2 different sets of models (each set containing 20 different models). Then for each of the test example, we have two expected influence scores corresponding to two different sets of models; we calculate the metrics (Pearson, Spearman and overlap between the $90^{th}$ percentile) between the two influence scores and report the average (across test examples) score in table~\ref{tab:marginal} {\sc expected}. We also compare it with a baseline, where we calculated the metrics using a pair of models (they are not expected influence); we repeat the baseline experiment with $50$ different pairs of models and report their average. Apart from leave one out and influence function all the other metrics are much more stable.

\section{Detecting Poisoned Training Examples as an Evaluation Metric}

\begin{figure*}
    \centering
    \subfloat[MNIST]{\includegraphics[scale=0.45]{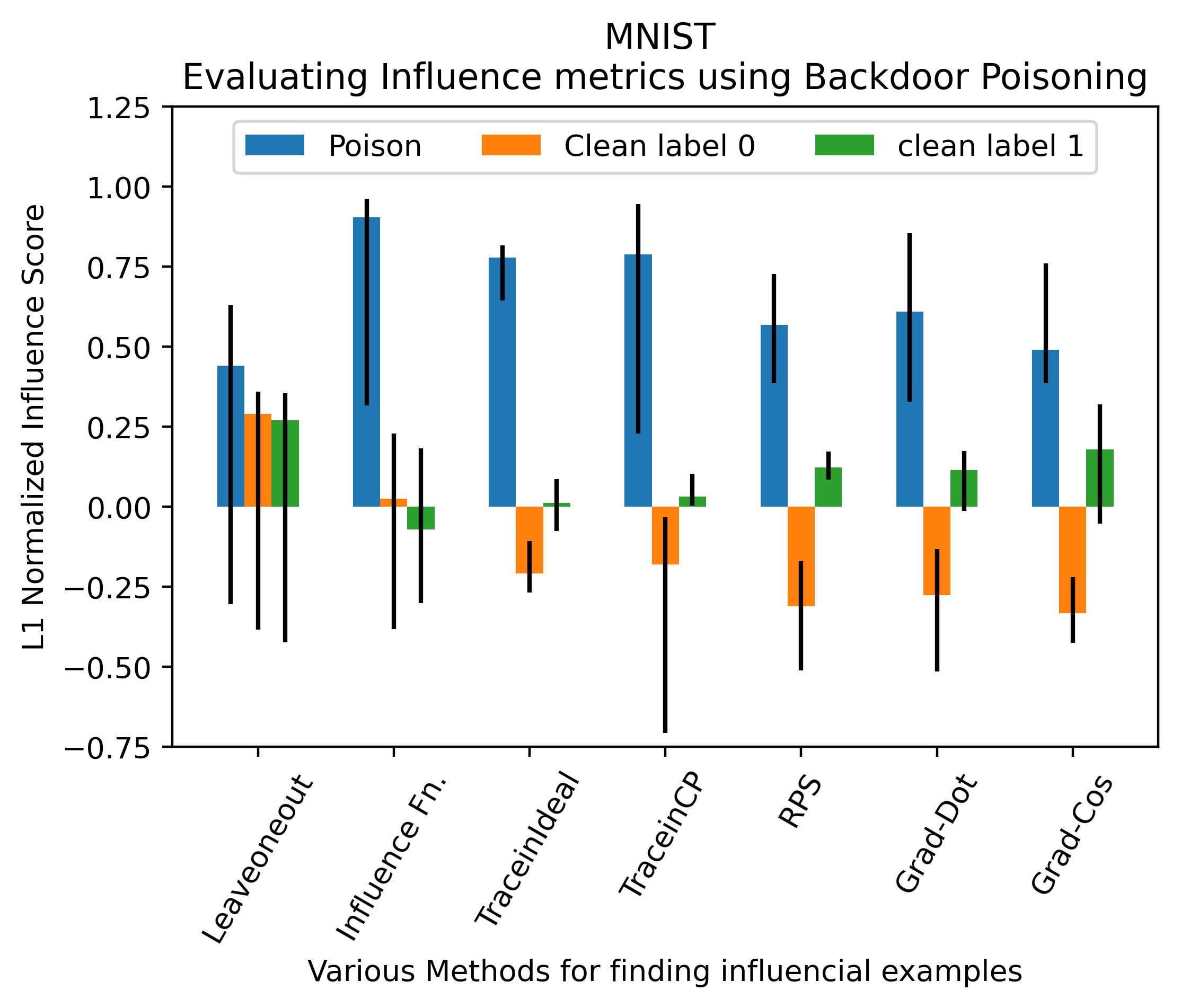}}
    \subfloat[CoMNIST]{\includegraphics[scale=0.45]{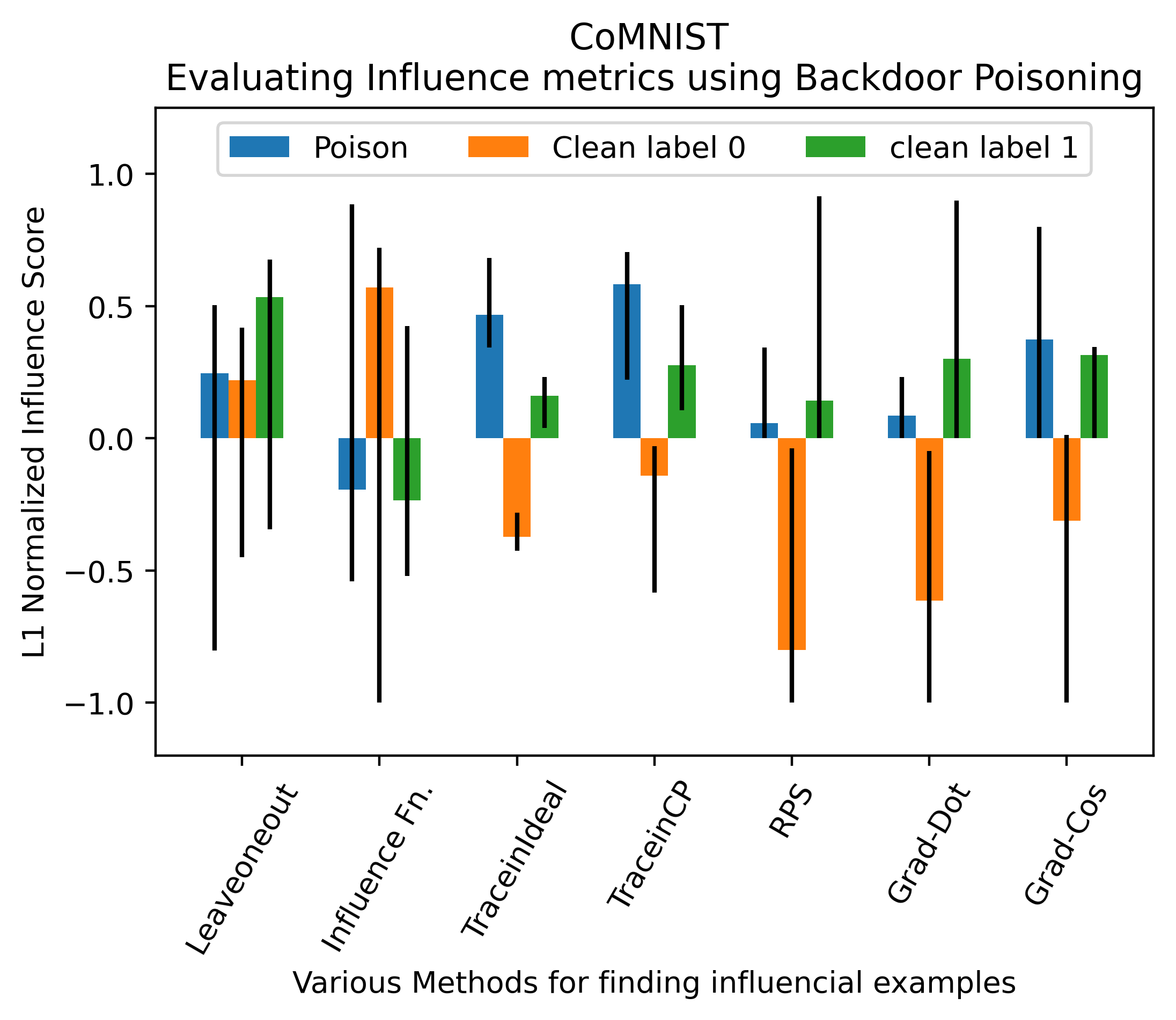}}
    \caption{\textbf{Detecting Poisoned training examples as an evaluation of influence score: } Intuitively, on poisoned test examples, poisoned training examples would be more influential than clean examples. For CoMNIST, we can see that, apart from Tracein and Grad-Cos other methods fail to identify poisoned training examples.}
    \label{fig:poison}
\end{figure*}




We propose a new approach to evaluate different influence methods that relies on backdoor poisoning attacks \citep{poison_svm,auditing_diff}. In the backdoor poisoning, we add a trigger or poison signature to a small portion of training examples from class A and relabel them as class B; then we train the model using this poisoned training data. 
Now, assume our model learns to predict a poisoned test example as class B. Then, the most influential training examples should be poisoned training examples. 
Hence, if the influence metric is meaningful, then it should detect that poisoned training examples are much more influential for correctly predicting poisoned test examples. 

For MNIST, we add a trigger to 100 training examples of class "0" and relabel them to class 1. Similarly, we take 100 test examples from class 0 and add the same trigger and relabel them to class 1. Note that {\em all our test examples are poisoned}. Similarly, for CoMNIST, we add trigger to examples of class A and relabel them to class B. We train 20 different models using this poisoned training data and calculate the expected influence scores (\S\ref{sec:expected_influence}) for each of the poisoned test data. For each of the test examples, we calculate the average expected influence of poisoned training examples, clean training examples of class 0 and 1, and then we average the scores across all the 100 test examples. Finally, we L1-normalize the three scores and plot them,  for each method, in Figure~\ref{fig:poison}. Again to re-illustrate that individual influence scores vary a lot, as described above, we calculate the metric using individual influence scores from each of the 20 models and report their minimum and maximum (across 20 models) as error bars in the plot. From Figure~\ref{fig:poison}, we can see that, on MNIST, all the expected influence scores are able to identify poisoned training examples. However, on Co-MNIST, only TraceinIdeal, TraceinCP and Grad-Cos were able to detect poisoned training examples.


\section{Conclusion}
This work begins with the observation (\S3) that five known methods (Influence Functions, Grad-Dot, Grad-Cos, TraceIn and Representer Point Selection) for finding influential examples are unstable, albeit some more than others: Influence functions \citep{pmlr-v70-koh17a} are extremely unstable, for example; Representer Point Selection \citep{yeh2018representer} less so.  
What is even more alarming is that leave-one-out influence estimates, which are often used as a gold standard for evaluating the methods for finding influential examples, are equally unstable. The gold standard for such methods is, in other words, unreliable. So is, moreover, recently proposed heuristics to evaluate these methods \citep{hanawa2021evaluation}. To remedy this, we revisit how influence is informally conceptualized, namely as something independent of the current model state and the rest of the dataset. We argue that influence should be thought of as relative to both things. 
Based on this, we propose to move from estimating influence to estimated {\em expected} influence. We show how this can be done in a simple, yet effective manner. Finally, we propose to evaluate methods for finding influential examples by their ability to detect poisoned training examples. We show that {\em expected}~TracIn performs best across the MNIST and CoMNIST datasets.  

\paragraph{Ethics} One can use methods for finding influential training instances for purposes that serve the common good, e.g., when using it to provide rationales for model decisions, improve system performance or to reduce social bias; but such methods can also be used for less noble purposes, e.g., censorship or to {\em increase}~social bias. We nevertheless think such methods are critical in making artificial intelligence transparent and democratic, and we hope this work, in providing a better foundation for employing and evaluating methods for finding influential examples, in general, can contribute to making these methods more useful and more broadly applicable. 




\bibliography{xai}

\end{document}


\maketitle

\appendix

 \section{Appendix}

We include the following equations for the change in loss on test examples under the original definition of leave-one-out (LOO) influence scores: 

\noindent \textbf{Original Equations: } 
\begin{align}
L_{init}(x_{test}) - L_{D}(x_{test}) = \sum_{i=1}^{N} I(x_i, x_{test}) \\
L_{init}(x_{test}) - L_{D/x_{train}}(x_{test}) = \sum_{\substack{i=1,\\ x_i\neq x_{train}}}^N  I(x_i, x_{test}) \\
I(x_{train}, x_{test}) = L_{D/x_{train}}(x_{test}) - L_{D}(x_{test}) \label{eqn:loo_original}
\end{align}

\noindent In the above, we argue that such influence scores should be conditioned on model state. This leads to the following corrected equations: 

\textbf{Corrected Equations (for batch size of 1)}
\begin{align}
\begin{split}
    L_{init}(x_{test}) - L_{D}(x_{test}) = \sum_{t=1}^{k} I(x_t, x_{test} | W_t ) \\ 
    \qquad + \sum_{t=k}^{T} I(x_t, x_{test} | W_t ) \\
\end{split}
\\[2ex]
\begin{split}
    L_{init}(x_{test}) - L_{D/x_{train}}(x_{test}) =  \sum_{t=1}^{k} I(x_t, x_{test} | W_t ) \\ 
    + \sum_{t=k}^{T^{new}} I(x_t, x_{test} | W_t^{new} ) \\
\end{split}
\\[2ex]
\begin{split}
 L_{D/x_{train}}(x_{test}) - L_{D}(x_{test}) = \sum_{t=k}^{T} I(x_t, x_{test} | W_t ) \\ 
 - \sum_{t=k}^{T^{new}} I(x_t, x_{test} | W_t^{new} ) \label{eqn:loo_corrected}
\end{split}
\end{align}
\noindent Where  $x_t, W_t$ is the training example and state at time step  $t$ respectively $T^{new}$ is the new total number of steps and  $W_t^{new}$ is the new state at time step $t$. 